\documentclass[11pt]{article}

\usepackage[final]{acl}
\usepackage{stroop}

\usepackage{times}
\usepackage{latexsym}
\usepackage{amsmath}

\usepackage[T1]{fontenc}

\usepackage[utf8]{inputenc}

\usepackage{microtype}

\usepackage{inconsolata}

\usepackage{graphicx}

\usepackage{framed}
\usepackage{fvextra}

\newcolumntype{G}{>{\color{gray}}c} %

\title{Asking a Language Model for Diverse Responses}

\author{
  Sergey Troshin$^*$\\
  University of Amsterdam \\
  \texttt{s.troshin@uva.nl}
  \And
  Irina Saparina$^*$ \\
  University of Edinburgh \\
  \texttt{i.saparina@sms.ed.ac.uk}
  \AND
  Antske Fokkens \\
  Vrije Universiteit Amsterdam \\
  \texttt{antske.fokkens@vu.nl}
  \And
  Vlad Niculae \\
  University of Amsterdam \\
  \texttt{v.niculae@uva.nl}
  \\
}

\begin{document}
\maketitle
\def\thefootnote{*}\footnotetext{These authors contributed equally to this work}\def\thefootnote{\arabic{footnote}}
\begin{abstract}
Large language models increasingly rely on explicit reasoning chains and can produce multiple plausible responses for a given context. We study the candidate sampler that produces the set of plausible responses contrasting the ancestral (parallel) sampling against two alternatives: enumeration, which asks the model to produce $n$ candidates in one pass, and iterative sampling, which proposes candidates sequentially while conditioning on the currently generated response set. Under matched budgets, we compare these samplers on quality, lexical and computation flow diversity, and efficiency. Our empirical results demonstrate that enumeration and iterative strategies result in higher diversity at comparable quality. Our findings highlight the potential of simple non-independent sampling strategies to improve response diversity without sacrificing generation quality.
\end{abstract}

\section{Introduction}
Large language models (LLMs) have shown strong performance across a wide range of applications \citep{openai2024gpt4technicalreport, deepseekai2025deepseekv3technicalreport}. In particular, the ability to generate explicit reasoning chains that guide planning and decision-making has become a cornerstone of recent progress \citep{chain_of_thought_elicit, yao2023tree, zhu2025soft, zhang-etal-2024-comprehensive-survey}. 
Many of these applications benefit from access to multiple plausible responses for a given context, including test-time control \citep{mudgal2024controlled, deng-raffel-2023-reward, troshin2025lowrankparametrizationrewardmodels}, majority voting or best-of-n \citep{stiennon2020learning, nakano2022webgptbrowserassistedquestionansweringhuman}, conformal generative modeling \citep{kladny2025conformal}, reasoning with diverse decoding paths \citep{wang2024mmlupro} and ambiguity resolution \citep{kobalczyk2025active, chen2025learning, saparina-lapata-2025-disambiguate}. 

A necessary component of these pipelines is a \textit{candidate sampler} that returns a set of $n$ responses in context. 
The candidates are commonly obtained by ancestral sampling
from the model distribution, or from variations such as temperature, top-p, top-k
\citep{Holtzman2020TheCuriousCase, basu2021mirostat, hewitt-etal-2022-truncation, nguyen2024turningheatminpsampling, vilnis_emb_sampling23}. 
Beyond being in some sense the natural approach, ancestral sampling also benefits from being simple to implement and readily 
parallelizable across devices, as each response is sampled independently of the others.
Nevertheless, ancestral sampling 
suffers from repetitions of high-probability sequences,
which motivated researchers to propose non-independent algorithms, including arithmetic sampling \citep{vilnis_emb_sampling23}, diverse, stochastic, and determinantal beam search modifications \citep{vijayakumar2018diversebeamsearchdecoding, kool2019stochastic, meister-etal-2021-determinantal}. 
These approaches, well-studied in the literature, are based on search-style algorithms on top of a language model's output probability, which still scores each sample separately, possibly with the help of a separate dissimilarity function.
In this work, we take a substantially different approach 
and ask whether we can use the standard LLM generation pipelines to enable efficient non-independent sampling, by processing multiple candidates at the same time.

In particular, we are interested in a candidate sampler that: 
\begin{enumerate}[label=(\roman*), leftmargin=*]
 \item produces high-quality samples;
 \item promotes response diversity;
 \item scales efficiently as the number of responses increases;
  \item is simple to use and relies on standard LLM decoding primitives.
 \end{enumerate}
 We compare the commonly used \textbf{parallel} sampling strategy (ancestral sampling) with two alternative sampling strategies, which we define as \textbf{enumeration} and \textbf{iterative} approaches, and study them from the perspective of quality, diversity, and efficiency.

Our main finding is that the enumeration and iterative strategies are simple and promising alternatives to the standard parallel approach. We find that our non-independent iterative and enumeration strategies result in higher lexical and computational flow diversity.
Such approaches can be seen in a way as upper-bound oracles to diverse generation, in the sense that they fully model the joint distribution over samples and are only limited by the instruction-following performance of the LLM.
Our implementation is released as open-source\footnote{\url{https://github.com/serjtroshin/ask4diversity}}.

\begin{figure*}[t]
\centering
\includegraphics[width=\linewidth]{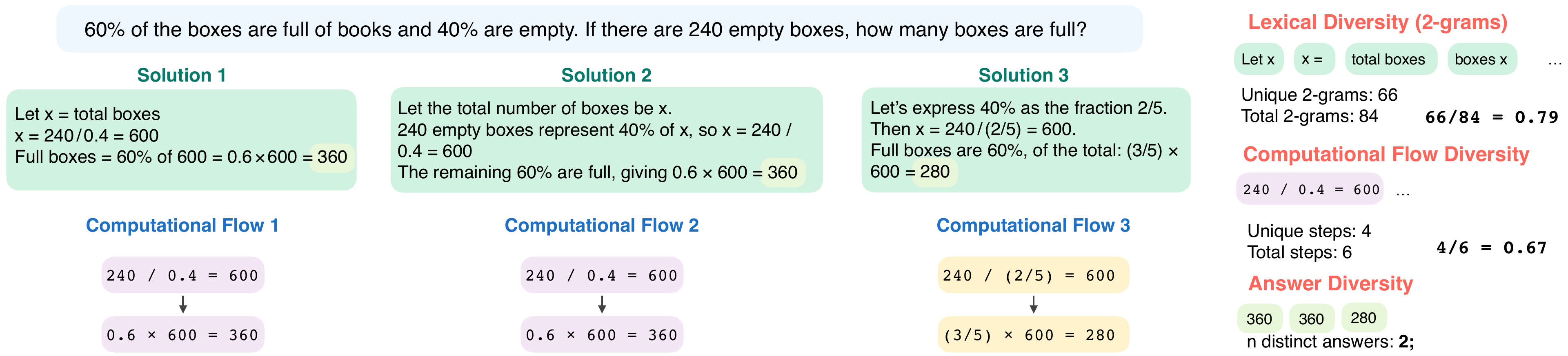}
\caption{Example of a math problem with three responses, their computation flows, and the resulting metrics: lexical, computational flow and answer diversity.}
\label{fig:metrics}
\end{figure*}

\section{Methodology}
We consider tasks for which there are multiple valid responses. In the context of this work, we consider a valid response to contain both a derivation and a final answer, so different derivations leading to the same answer are valid responses.
Given a model $p_\theta$ and a prompt $c$, our goal is to produce a set $\mathcal{S}=\{y^{(1)},\dots,y^{(n)}\}$ of $n$ responses. We keep all decoding hyperparameters fixed across methods and vary only the sampling protocol.

\subsection{Sampling Strategies}
\paragraph{Parallel sampling.} We sample $n$ times independently with different random seeds; samples do not condition on one another:
\begin{equation}
    y^{(i)} \sim p_\theta(\cdot \mid c;)\quad\text{for }i=1..n
\end{equation}

\paragraph{Enumeration sampling.} 
We prompt the model to generate multiple different outputs in one pass; later outputs condition on earlier ones:
\begin{equation}
    y^{(k)} \sim \prod_{i=1}^{k} p_\theta\left(y^{(i)} \mid c,\, y^{(1:i-1)}\right).
\end{equation}
The number of desired samples is not specified in the prompt, but rather implicitly predicted.
To the best of our knowledge, the enumeration approach has not been studied in the literature. However, due to its simplicity, we speculate it it is used in practice, for example, \citet{ilia-aziz-2024-predict} prompt ChatGPT \citep{openai2022chatgpt} to enumerate $40$ responses in context as a complementary strategy to ancestral sampling; \citet{saparina2024ambrosia} prompt models to enumerate all possible interpretations of ambiguous questions.

\paragraph{Iterative sampling.} We generate one candidate at a time, and we re-prompt the model to extend an already generated list of responses with a new response. Namely, for $k=1$, we generate as:
\begin{equation}
    y^{(1)} \sim p_\theta(\cdot \mid c),
\end{equation}
and for $k>1$, we pass the generated solutions:
\begin{equation}
    y^{(k)} \sim p_\theta\left(\cdot \mid c(y^{(1)},\cdots,y^{(k-1)})\right).
\end{equation}
In practice, the conditioning is achieved with a templated prompt; refer to Appendix~\ref{app:prompts} for the specific prompts used for all strategies.

\section{Experimental Setup}
We evaluate on GSM8K \citep{cobbe2021gsm8k}, a grade school math problem-solving benchmark. Each problem has a single gold answer, but multiple valid solutions may lead to it. Therefore, a candidate is \(y^{(i)}=(r^{(i)},a^{(i)})\), with \(r^{(i)}\) the solution (reasoning) and \(a^{(i)}\) the final extracted answer.

\subsection{Models}
In our work, we rely on the Qwen3 family of models \citep{yang2025qwen3technicalreport}, chosen for their high reasoning performance, diverse range of model sizes. In our preliminary investigation, we observe that Qwen3 models are able to follow our zero-shot instructions, and they show high accuracy in following the required output format. For our experiments, we use \texttt{Qwen3-\{4B,8B,14B\}} models with thinking generation mode on; and we use \texttt{Qwen3-4B-\{Instruct/Thinking\}-2507} released solely for non-thinking/thinking use-cases.

We use the hyperparameters suggested by the model developers: \texttt{temperature} $= 0.6$, \texttt{top-}$k$ $= 20$, \texttt{top-}$p$ $= 0.95$, \texttt{repetition\_penalty} $=1.0$.

\subsection{Metrics}

\paragraph{Quality.}
We define the quality metrics as the average accuracy over response sets given a golden answer for a problem. 
We calculate the accuracy of a response set by taking the minimum, mean, and maximum statistics over the answers within the set and averaging these statistics over the dataset.

\paragraph{Lexical diversity.}
We follow \citet{li-etal-2016-diversity} and report \textbf{averaged distinct $N$-gram diversity} metric as the proportion of distinct $N$-grams in the set of responses relative to the total number of $N$-grams.

\paragraph{Computation flow diversity.}
To complement the lexical diversity metric, we extract a computation flow of each solution by mapping it to sequences of normalized arithmetic steps (\eg, ``Janet sells 9 eggs at \$2 each, which gives 18'' maps to $9 \times 2 = 18$). We obtain flows with a one-shot prompt to Qwen-3-32B (see Appendix~\ref{app:comp_flow_prompt}). We report the proportion of unique steps relative to the total number of steps in the set. To compute this metric, we estimate the distinct 1-grams over the simple arithmetical steps, namely $9 \times 2 = 18$ is considered to be a single 1-gram. This approach can collapse steps that are arithmetically identical but occur in different parts of a solution; however, we found this to be rare in our experiments. If needed, repeated occurrences can be distinguished by indexing them within a flow (e.g., (1) $9 \times 2 = 18$, (2) $9 \times 2 = 18$).

\paragraph{Final answer variability.}
For some applications, it might be useful to have samples with different answers (\eg to have both positive and negative demonstrations), and we measure the answer variability as the number of unique answers among the response set. For GSM8K, high answer variability means that some answers are parsed as incorrect.

Figure~\ref{fig:metrics} illustrates an input math problem, three different responses, the corresponding computation flows, and the resulting metrics. The first and second responses differ in phrasing, but follow the same computation; the third differs in wording and computation but yields an incorrect result.

\section{Results}
\subsection{Quality and Diversity}
In \Cref{tab:main_results}, we report the evaluation results on the GSM8K dataset. 
\paragraph{Parsing the solutions.}
We parse the responses from the generated outputs by searching for the required solution tags, \ie, \texttt{<Solution>...</Solution>}. For the \textit{parallel} and \textit{iteration} strategies, we obtain more than $4$ successfully parsed responses on average (out of $5$ required). For the \textit{enumeration} strategy, we do not specify the required number of responses and obtain between 2 and 4 parsed responses on average. Overall, Qwen3 models demonstrate a satisfactory ability to follow our instructions for output formatting.

\paragraph{Diversity of the responses.}
We observe that in all cases the diversity of samples from the \textit{parallel} strategy is lower compared to the diversity of the two non-independent strategies, both for the lexical and computational flow diversity. We observe that often higher lexical diversity does not imply higher compute diversity, and we think these metrics can provide complementary signals to the developers.

\paragraph{Quality of the answers.}
In most cases, our models demonstrate good zero-shot task performance with an accuracy of around $90$\%. Parallel sampling shows the most stable high quality (lowest quality variation), probably because it is the most standard approach, and it is easier for a language model to adapt to the corresponding prompt requirements.

\begin{table*}[ht]
    \centering
    \small
    \begin{tabular}{l l c G c G c c c}
        \toprule
       &                 &   \# Parsed  & Min & Mean &  Max    & Lexical   &  Compute          & \# Distinct   \\
        Model & Strategy &  Solutions  & Quality & Quality & Quality& Diversity &  Diversity        & Answers \\     
\midrule
Qwen3-4B             & parallel     & 4.77 & 0.86 &  0.91  & 0.95 & 42.8    &   33.1   & 1.13    \\
                    & enumeration  & 3.90 & 0.88 &  0.90  & 0.91 & 68.1    &   56.1    & 1.04    \\
                    & iteration    & 4.75 & 0.83 &  0.87  & 0.90 & 61.8    &   60.0    & 1.07    \\ \addlinespace
Qwen3-8B             & parallel     & 4.23 & 0.89 &  0.91  & 0.93 & 44.5    &   34.7   & 1.06        \\
                    & enumeration  & 2.81 & 0.89 &  0.90  & 0.91 & 73.1    &   64.1    & 1.03      \\
                    & iteration    & 4.87 & 0.89 &  0.91  & 0.92 & 63.4    &   79.8    & 1.03         \\  \addlinespace
Qwen3-14B            & parallel     & 4.90 & 0.92 &  0.94  & 0.96 & 38.4    &   31.5   & 1.05     \\
                    & enumeration  & 3.58 & 0.90 &  0.92  & 0.94 & 70.2    &   57.1    & 1.05     \\
                    & iteration    & 4.96 & 0.60 &  0.73  & 0.83 & 70.1    &   59.3    & 1.25     \\  \addlinespace
Qwen3-4B-Instruct    & parallel     & 4.98 & 0.88 &  0.92  & 0.94 & 33.1    &   47.7   & 1.10   \\
                    & enumeration  & 3.09 & 0.88 &  0.90  & 0.91 & 72.8    &    61.2   & 1.04  \\
                    & iteration    & 5.00 & 0.86 &  0.89  & 0.90 & 60.3    &    55.6   & 1.08  \\  \addlinespace
Qwen3-4B-Thinking    & parallel     & 4.67 & 0.81 &  0.89  & 0.94 & 47.9    &    30.7  & 1.24  \\
                    & enumeration  & 2.27 & 0.64 &  0.73  & 0.79 & 66.2    &    64.5   & 1.19  \\
                    & iteration    & 4.17 & 0.78 &  0.87  & 0.92 & 68.0    &    62.0   & 1.16  \\
\bottomrule
    \end{tabular}
    \caption{Main results for \textit{parallel}, \textit{enumeration}, and \textit{iteration} sampling strategies. For enumeration, we let the model decide the number of solutions, for parallel and iteration, we expect $5$ solutions, and report the average number of parsed solutions. Min and max quality denote the average minimum and maximum accuracy over the response sets. \# distinct answers denote the average number of distinct answers %
    among the set of parsed responses.}
    \label{tab:main_results}
\end{table*}

\paragraph{Variability of the answers.}
Additionally, we report the answer variability and the average minimum and maximum accuracy over the responses. We observe that overall models exhibit low answer variability with less than $1.3$ distinct answers on average. Enumeration strategy results in the highest quality difference (i.e., the gap between maximum and  minimum accuracy), while the parallel and iteration are on par with each other. We note that under diversity requirements, we do not expect a model to always produce a parsable or even correct answer, and part of the quality loss can be attributed to answer parser failures.

\subsection{Compute Efficiency}
An important question when developing the sampling strategies is to understand how efficient it is to generate the set of $n$ responses. We distinguish the total number of generation calls that we need to do in order to generate $n$ responses, and the support for parallelization. We compare the three strategies \wrt the compute they require.

From the perspective of parallel-time computation, the \textit{parallel} approach is most time-efficient by design, and this sort of parallelization is well optimized and supported in LLM codebases, but its time efficiency is tied to the access to parallel computation (\eg, a multi-GPU setup). As we observe from the diversity results, the independence assumption results in lower diversity (a higher degree of repetitions).

Both enumeration and iteration are most suited for single-GPU generation. 
For \textit{enumeration}, we need a single call to the model to enumerate the generations in the response; in thinking mode, the model shares the computation to produce $n$ responses: it generates a single thinking chain first, and then it enumerates the responses. 
A limitation of this strategy is that this approach requires a larger context length to produce multiple responses in one go, which in turn slows down the decoding for the standard quadratic-time attention implementation.

For \textit{iteration}, we need $n$ full sequential calls: the generated responses are reused, but not any other internals.
Iteration is less time-efficient than enumeration, since the former
requires multiple sequential generation calls; on the other hand, iteration sampling allows for easy and more explicit control of the number of responses, and may be more compatible with other probabilistic modeling strategies for subset selection without sacrificing the expressiveness of enumeration sampling.

 The main difference between parallel and the two serial approaches (enumeration and iteration) is the degree to which information is shared and efficiently reused across the set when generating responses. 
We see promise in further study of information conditioning and compression, specifically, quantifying the extent of this sharing and reuse. %
 In particular, the enumeration strategy can potentially approach the efficiency of a single parallel call while processing the responses quasi-independently, which in turn affects the diversity of the responses.

\section{Conclusion}

We study the problem of generating a diverse set of responses. We propose two non-independent approaches for sampling responses from a language model, namely enumeration and iteration strategies, and compare them against parallel algorithms based on ancestral sampling. On  GSM8k, we find that our non-independent approaches can provide higher diversity of the samples, while maintaining simplicity and overall quality of the generations. Compute efficiency analysis shows that enumeration and iteration are well-suited to a single GPU and can reduce redundancy without specialized search machinery. We hope our work will motivate further investigation of simple non-independent strategies for diverse candidate sampling.

\section{Limitations}
One of the main limitations of our work is a narrow evaluation scope. We focus on a single dataset with verifiable rewards and a room for diversity of answers and reasoning chains. Future work can evaluate these methods on tasks that inherently benefit from diverse generations, such as creative writing, code generation, or ambiguous question answering.
We do not compare the results to established diverse decoding methods such as beam search variants, as we limit our scope to sampling from the model output distribution rather than modifying it through specialized decoding algorithms. \citet{ippolito-etal-2019-comparison} provide an extensive survey and evaluation methodology for the established methods.

\section*{Acknowledgments}
This work is part of the UTTER project, supported by the European Union's Horizon Europe research and innovation programme via grant agreement 101070631.
This work is also supported by project VI.Veni.212.228 of the research program
`Veni', which is financed by the Dutch Research Council (NWO); and is part of
‘Hybrid Intelligence: augmenting human intellect’
(https://hybrid-intelligence-centre.nl) with project number 024.004.022 of the
research program `Gravitation' which is (partly) financed by the Dutch Research
Council (NWO). 

\bibliography{anthology,custom}

\begin{thebibliography}{31}
\providecommand{\natexlab}[1]{#1}

\bibitem[{Basu et~al.(2021)Basu, Ramachandran, Keskar, and Varshney}]{basu2021mirostat}
Sourya Basu, Govardana~Sachitanandam Ramachandran, Nitish~Shirish Keskar, and Lav~R. Varshney. 2021.
\newblock \href {https://openreview.net/forum?id=W1G1JZEIy5_} {Mirostat: A neural text decoding algorithm that directly controls perplexity}.
\newblock In \emph{ICLR}.

\bibitem[{Chen et~al.(2025)Chen, Sun, Pfister, and Arik}]{chen2025learning}
Maximillian Chen, Ruoxi Sun, Tomas Pfister, and Sercan~O Arik. 2025.
\newblock \href {https://openreview.net/forum?id=SIE6VFps9x} {Learning to clarify: Multi-turn conversations with action-based contrastive self-training}.
\newblock In \emph{The Thirteenth International Conference on Learning Representations}.

\bibitem[{Cobbe et~al.(2021)Cobbe, Kosaraju, Bavarian, Chen, Jun, Kaiser, Plappert, Tworek, Hilton, Nakano, Hesse, and Schulman}]{cobbe2021gsm8k}
Karl Cobbe, Vineet Kosaraju, Mohammad Bavarian, Mark Chen, Heewoo Jun, Lukasz Kaiser, Matthias Plappert, Jerry Tworek, Jacob Hilton, Reiichiro Nakano, Christopher Hesse, and John Schulman. 2021.
\newblock Training verifiers to solve math word problems.
\newblock \emph{arXiv preprint arXiv:2110.14168}.

\bibitem[{DeepSeek-AI et~al.(2025)DeepSeek-AI, Liu, Feng, Xue, Wang, Wu, Lu, Zhao, Deng, Zhang, Ruan, Dai, Guo, Yang, Chen, Ji, Li, Lin, Dai, Luo, Hao, Chen, Li, Zhang, Bao, Xu, Wang, Zhang, Ding, Xin, Gao, Li, Qu, and et~al.}]{deepseekai2025deepseekv3technicalreport}
DeepSeek-AI, Aixin Liu, Bei Feng, Bing Xue, Bingxuan Wang, Bochao Wu, Chengda Lu, Chenggang Zhao, Chengqi Deng, Chenyu Zhang, Chong Ruan, Damai Dai, Daya Guo, Dejian Yang, Deli Chen, Dongjie Ji, Erhang Li, Fangyun Lin, Fucong Dai, and 15 others. 2025.
\newblock \href {https://arxiv.org/abs/2412.19437} {Deepseek-v3 technical report}.
\newblock \emph{Preprint}, arXiv:2412.19437.

\bibitem[{Deng and Raffel(2023)}]{deng-raffel-2023-reward}
Haikang Deng and Colin Raffel. 2023.
\newblock \href {https://doi.org/10.18653/v1/2023.emnlp-main.721} {Reward-augmented decoding: Efficient controlled text generation with a unidirectional reward model}.
\newblock In \emph{Proceedings of the 2023 Conference on Empirical Methods in Natural Language Processing}, pages 11781--11791, Singapore. Association for Computational Linguistics.

\bibitem[{Hewitt et~al.(2022)Hewitt, Manning, and Liang}]{hewitt-etal-2022-truncation}
John Hewitt, Christopher Manning, and Percy Liang. 2022.
\newblock \href {https://doi.org/10.18653/v1/2022.findings-emnlp.249} {Truncation sampling as language model desmoothing}.
\newblock In \emph{Findings of the Association for Computational Linguistics: EMNLP 2022}, pages 3414--3427, Abu Dhabi, United Arab Emirates. Association for Computational Linguistics.

\bibitem[{Holtzman et~al.(2020)Holtzman, Buys, Du, Forbes, and Choi}]{Holtzman2020TheCuriousCase}
Ari Holtzman, Jan Buys, Li~Du, Maxwell Forbes, and Yejin Choi. 2020.
\newblock \href {https://openreview.net/forum?id=rygGQyrFvH} {The curious case of neural text degeneration}.
\newblock In \emph{ICLR}.

\bibitem[{Ilia and Aziz(2024)}]{ilia-aziz-2024-predict}
Evgenia Ilia and Wilker Aziz. 2024.
\newblock \href {https://aclanthology.org/2024.eacl-short.22/} {Predict the next word: {\ensuremath{<}}humans exhibit uncertainty in this task and language models {\_}{\_}{\_}{\_}{\_}{\ensuremath{>}}}.
\newblock In \emph{Proceedings of the 18th Conference of the European Chapter of the Association for Computational Linguistics (Volume 2: Short Papers)}, pages 234--255, St. Julian{'}s, Malta. Association for Computational Linguistics.

\bibitem[{Ippolito et~al.(2019)Ippolito, Kriz, Sedoc, Kustikova, and Callison-Burch}]{ippolito-etal-2019-comparison}
Daphne Ippolito, Reno Kriz, Jo{\~a}o Sedoc, Maria Kustikova, and Chris Callison-Burch. 2019.
\newblock \href {https://doi.org/10.18653/v1/P19-1365} {Comparison of diverse decoding methods from conditional language models}.
\newblock In \emph{Proceedings of the 57th Annual Meeting of the Association for Computational Linguistics}, pages 3752--3762, Florence, Italy. Association for Computational Linguistics.

\bibitem[{Kladny et~al.(2025)Kladny, Sch{\"o}lkopf, and Muehlebach}]{kladny2025conformal}
Klaus-Rudolf Kladny, Bernhard Sch{\"o}lkopf, and Michael Muehlebach. 2025.
\newblock \href {https://openreview.net/forum?id=1i6lkavJ94} {Conformal generative modeling with improved sample efficiency through sequential greedy filtering}.
\newblock In \emph{The Thirteenth International Conference on Learning Representations}.

\bibitem[{Kobalczyk et~al.(2025)Kobalczyk, Astorga, Liu, and van~der Schaar}]{kobalczyk2025active}
Kasia Kobalczyk, Nicol{\'a}s Astorga, Tennison Liu, and Mihaela van~der Schaar. 2025.
\newblock \href {https://openreview.net/forum?id=JAMxRSXLFz} {Active task disambiguation with {LLM}s}.
\newblock In \emph{The Thirteenth International Conference on Learning Representations}.

\bibitem[{Kool et~al.(2019)Kool, van Hoof, and Welling}]{kool2019stochastic}
Wouter Kool, Herke van Hoof, and Max Welling. 2019.
\newblock Stochastic beams and where to find them: The gumbel-top-k trick for sampling sequences without replacement.
\newblock In \emph{ICML}.

\bibitem[{Li et~al.(2016)Li, Galley, Brockett, Gao, and Dolan}]{li-etal-2016-diversity}
Jiwei Li, Michel Galley, Chris Brockett, Jianfeng Gao, and Bill Dolan. 2016.
\newblock \href {https://doi.org/10.18653/v1/N16-1014} {A diversity-promoting objective function for neural conversation models}.
\newblock In \emph{Proceedings of the 2016 Conference of the North {A}merican Chapter of the Association for Computational Linguistics: Human Language Technologies}, pages 110--119, San Diego, California. Association for Computational Linguistics.

\bibitem[{Meister et~al.(2021)Meister, Forster, and Cotterell}]{meister-etal-2021-determinantal}
Clara Meister, Martina Forster, and Ryan Cotterell. 2021.
\newblock \href {https://doi.org/10.18653/v1/2021.acl-long.512} {Determinantal beam search}.
\newblock In \emph{Proceedings of the 59th Annual Meeting of the Association for Computational Linguistics and the 11th International Joint Conference on Natural Language Processing (Volume 1: Long Papers)}, pages 6551--6562, Online. Association for Computational Linguistics.

\bibitem[{Minh et~al.(2025)Minh, Baker, Neo, Roush, Kirsch, and Shwartz-Ziv}]{nguyen2024turningheatminpsampling}
Nguyen~Nhat Minh, Andrew Baker, Clement Neo, Allen~G Roush, Andreas Kirsch, and Ravid Shwartz-Ziv. 2025.
\newblock \href {https://openreview.net/forum?id=FBkpCyujtS} {Turning up the heat: Min-p sampling for creative and coherent {LLM} outputs}.
\newblock In \emph{ICLR}.

\bibitem[{Mudgal et~al.(2024)Mudgal, Lee, Ganapathy, Li, Wang, Huang, Chen, Cheng, Collins, Strohman, Chen, Beutel, and Beirami}]{mudgal2024controlled}
Sidharth Mudgal, Jong Lee, Harish Ganapathy, YaGuang Li, Tao Wang, Yanping Huang, Zhifeng Chen, Heng-Tze Cheng, Michael Collins, Trevor Strohman, Jilin Chen, Alex Beutel, and Ahmad Beirami. 2024.
\newblock \href {https://openreview.net/forum?id=bVIcZb7Qa0} {Controlled decoding from language models}.
\newblock In \emph{ICML}.

\bibitem[{Nakano et~al.(2022)Nakano, Hilton, Balaji, Wu, Ouyang, Kim, Hesse, Jain, Kosaraju, Saunders, Jiang, Cobbe, Eloundou, Krueger, Button, Knight, Chess, and Schulman}]{nakano2022webgptbrowserassistedquestionansweringhuman}
Reiichiro Nakano, Jacob Hilton, Suchir Balaji, Jeff Wu, Long Ouyang, Christina Kim, Christopher Hesse, Shantanu Jain, Vineet Kosaraju, William Saunders, Xu~Jiang, Karl Cobbe, Tyna Eloundou, Gretchen Krueger, Kevin Button, Matthew Knight, Benjamin Chess, and John Schulman. 2022.
\newblock \href {https://arxiv.org/abs/2112.09332} {Webgpt: Browser-assisted question-answering with human feedback}.
\newblock \emph{Preprint}, arXiv:2112.09332.

\bibitem[{OpenAI(2022)}]{openai2022chatgpt}
OpenAI. 2022.
\newblock Introducing {ChatGPT}.
\newblock \url{https://openai.com/blog/chatgpt}.
\newblock Accessed August 15, 2025.

\bibitem[{OpenAI et~al.(2024)OpenAI, Achiam, Adler, Agarwal, Ahmad, Akkaya, Aleman, Almeida, Altenschmidt, Altman, Anadkat, Avila, Babuschkin, Balaji, Balcom, Baltescu, Bao, Bavarian, Belgum, Bello, Berdine, Bernadett-Shapiro, Berner, Bogdonoff, Boiko, Boyd, Brakman, Brockman, Brooks, Brundage, Button, Cai, Campbell, Cann, Carey, Carlson, Carmichael, Chan, Chang, Chantzis, Chen, Chen, and et~al.}]{openai2024gpt4technicalreport}
OpenAI, Josh Achiam, Steven Adler, Sandhini Agarwal, Lama Ahmad, Ilge Akkaya, Florencia~Leoni Aleman, Diogo Almeida, Janko Altenschmidt, Sam Altman, Shyamal Anadkat, Red Avila, Igor Babuschkin, Suchir Balaji, Valerie Balcom, Paul Baltescu, Haiming Bao, Mohammad Bavarian, Jeff Belgum, and 24 others. 2024.
\newblock \href {https://arxiv.org/abs/2303.08774} {Gpt-4 technical report}.
\newblock \emph{Preprint}, arXiv:2303.08774.

\bibitem[{Saparina and Lapata(2024)}]{saparina2024ambrosia}
Irina Saparina and Mirella Lapata. 2024.
\newblock \href {https://proceedings.neurips.cc/paper_files/paper/2024/file/a4c942a8405cc910f0a833d28d2573cc-Paper-Datasets_and_Benchmarks_Track.pdf} {Ambrosia: A benchmark for parsing ambiguous questions into database queries}.
\newblock In \emph{Advances in Neural Information Processing Systems}, volume~37, pages 90600--90628. Curran Associates, Inc.

\bibitem[{Saparina and Lapata(2025)}]{saparina-lapata-2025-disambiguate}
Irina Saparina and Mirella Lapata. 2025.
\newblock \href {https://doi.org/10.18653/v1/2025.findings-acl.863} {Disambiguate first, parse later: Generating interpretations for ambiguity resolution in semantic parsing}.
\newblock In \emph{Findings of the Association for Computational Linguistics: ACL 2025}, pages 16825--16839, Vienna, Austria. Association for Computational Linguistics.

\bibitem[{Stiennon et~al.(2020)Stiennon, Ouyang, Wu, Ziegler, Lowe, Voss, Radford, Amodei, and Christiano}]{stiennon2020learning}
Nisan Stiennon, Long Ouyang, Jeff Wu, Daniel~M. Ziegler, Ryan Lowe, Chelsea Voss, Alec Radford, Dario Amodei, and Paul Christiano. 2020.
\newblock \href {https://proceedings.neurips.cc/paper/2020/hash/1f89885d556929e98d3ef9b86448f951-Abstract.html} {Learning to summarize with human feedback}.
\newblock In \emph{NeurIPS}.

\bibitem[{Troshin et~al.(2025)Troshin, Niculae, and Fokkens}]{troshin2025lowrankparametrizationrewardmodels}
Sergey Troshin, Vlad Niculae, and Antske Fokkens. 2025.
\newblock \href {https://arxiv.org/abs/2407.04615} {On the low-rank parametrization of reward models for controlled language generation}.
\newblock In \emph{TMLR}.

\bibitem[{Vijayakumar et~al.(2018)Vijayakumar, Cogswell, Selvaraju, Sun, Lee, Crandall, and Batra}]{vijayakumar2018diversebeamsearchdecoding}
Ashwin~K. Vijayakumar, Michael Cogswell, Ramprasaath~R. Selvaraju, Qing Sun, Stefan Lee, David Crandall, and Dhruv Batra. 2018.
\newblock \href {https://doi.org/10.1609/aaai.v32i1.12340} {Diverse beam search for improved description of complex scenes}.
\newblock In \emph{AAAI}.

\bibitem[{Vilnis et~al.(2023)Vilnis, Zemlyanskiy, Murray, Passos, and Sanghai}]{vilnis_emb_sampling23}
Luke Vilnis, Yury Zemlyanskiy, Patrick Murray, Alexandre Passos, and Sumit Sanghai. 2023.
\newblock Arithmetic sampling: parallel diverse decoding for large language models.
\newblock In \emph{ICML}.

\bibitem[{Wang et~al.(2024)Wang, Ma, Zhang, Ni, Chandra, Guo, Ren, Arulraj, He, Jiang, Li, Ku, Wang, Zhuang, Fan, Yue, and Chen}]{wang2024mmlupro}
Yubo Wang, Xueguang Ma, Ge~Zhang, Yuansheng Ni, Abhranil Chandra, Shiguang Guo, Weiming Ren, Aaran Arulraj, Xuan He, Ziyan Jiang, Tianle Li, Max Ku, Kai Wang, Alex Zhuang, Rongqi Fan, Xiang Yue, and Wenhu Chen. 2024.
\newblock \href {https://openreview.net/forum?id=y10DM6R2r3} {{MMLU}-pro: A more robust and challenging multi-task language understanding benchmark}.
\newblock In \emph{The Thirty-eight Conference on Neural Information Processing Systems Datasets and Benchmarks Track}.

\bibitem[{Wei et~al.(2022)Wei, Wang, Schuurmans, Bosma, Ichter, Xia, Chi, Le, and Zhou}]{chain_of_thought_elicit}
Jason Wei, Xuezhi Wang, Dale Schuurmans, Maarten Bosma, Brian Ichter, Fei Xia, Ed~H. Chi, Quoc~V. Le, and Denny Zhou. 2022.
\newblock Chain-of-thought prompting elicits reasoning in large language models.
\newblock In \emph{NeurIPS}.

\bibitem[{Yang et~al.(2025)Yang, Li, Yang, Zhang, Hui, Zheng, Yu, Gao, Huang, Lv, Zheng, Liu, Zhou, Huang, Hu, Ge, Wei, Lin, Tang, Yang, Tu, Zhang, Yang, Yang, Zhou, Zhou, Lin, Dang, Bao, Yang, Yu, Deng, Li, Xue, Li, Zhang, Wang, Zhu, Men, Gao, Liu, Luo, Li, Tang, Yin, Ren, Wang, Zhang, Ren, Fan, Su, Zhang, Zhang, Wan, Liu, Wang, Cui, Zhang, Zhou, and Qiu}]{yang2025qwen3technicalreport}
An~Yang, Anfeng Li, Baosong Yang, Beichen Zhang, Binyuan Hui, Bo~Zheng, Bowen Yu, Chang Gao, Chengen Huang, Chenxu Lv, Chujie Zheng, Dayiheng Liu, Fan Zhou, Fei Huang, Feng Hu, Hao Ge, Haoran Wei, Huan Lin, Jialong Tang, and 41 others. 2025.
\newblock \href {https://arxiv.org/abs/2505.09388} {Qwen3 technical report}.
\newblock \emph{Preprint}, arXiv:2505.09388.

\bibitem[{Yao et~al.(2023)Yao, Yu, Zhao, Shafran, Griffiths, Cao, and Narasimhan}]{yao2023tree}
Shunyu Yao, Dian Yu, Jeffrey Zhao, Izhak Shafran, Thomas~L. Griffiths, Yuan Cao, and Karthik~R Narasimhan. 2023.
\newblock \href {https://openreview.net/forum?id=5Xc1ecxO1h} {Tree of thoughts: Deliberate problem solving with large language models}.
\newblock In \emph{NeurIPS}.

\bibitem[{Zhang et~al.(2024)Zhang, Chen, Jin, Wang, Ji, Wang, and Han}]{zhang-etal-2024-comprehensive-survey}
Yu~Zhang, Xiusi Chen, Bowen Jin, Sheng Wang, Shuiwang Ji, Wei Wang, and Jiawei Han. 2024.
\newblock \href {https://doi.org/10.18653/v1/2024.emnlp-main.498} {A comprehensive survey of scientific large language models and their applications in scientific discovery}.
\newblock In \emph{Proceedings of the 2024 Conference on Empirical Methods in Natural Language Processing}, pages 8783--8817, Miami, Florida, USA. Association for Computational Linguistics.

\bibitem[{Zhu et~al.(2025)Zhu, Zhao, Yan, He, Chen, and Gui}]{zhu2025soft}
Qinglin Zhu, Runcong Zhao, Hanqi Yan, Yulan He, Yudong Chen, and Lin Gui. 2025.
\newblock \href {https://openreview.net/forum?id=4gWE7CMOlH} {Soft reasoning: Navigating solution spaces in large language models through controlled embedding exploration}.
\newblock In \emph{ICML}.

\end{thebibliography}

\appendix

\section{Prompts}\label{app:prompts}
\textbf{Prompt for enumeration sampling.}

\begin{framed}
\vspace{-3mm}
\begin{Verbatim}[breaklines=true, 
                 breakanywhere=true,
                 breaksymbolright={},
                 breaksymbolsepleft=0pt,
                 breaksymbolindent=0pt,
                 breaksymbolleft={},
                 fontsize=\small]
Given the following problem, reason through it and provide multiple different solutions:

Problem: {question}

Use exactly this format (no extra text):
<Solution 1> [Your reasoning should go here] The answer is [Answer 1]. </Solution 1>
...
<Solution N> [Your reasoning should go here] The answer is [Answer N]. </Solution N>
\end{Verbatim}
\vspace{-3mm}
\end{framed}

\textbf{Prompt for parallel sampling.}
\begin{framed}
\vspace{-3mm}
\begin{Verbatim}[breaklines=true, 
                 breakanywhere=true,
                 breaksymbolright={},
                 breaksymbolsepleft=0pt,
                 breaksymbolindent=0pt,
                 breaksymbolleft={},
                 fontsize=\small]
Given the following problem, reason through it and provide a solution:

Problem: {question}

You must wrap your reasoning and answer into <Solution> ...reasoning here... 'The answer is [numerical value].'</Solution> format.
\end{Verbatim}
\vspace{-3mm}
\end{framed}

\textbf{Prompt for iterative sampling.}

\begin{framed}
\vspace{-3mm}
\begin{Verbatim}[breaklines=true, 
                 breakanywhere=true,
                 breaksymbolright={},
                 breaksymbolsepleft=0pt,
                 breaksymbolindent=0pt,
                 breaksymbolleft={},
                 fontsize=\small]
Given a problem and a set of solutions, reason through it and provide a new solution. The new solution may result in the same answer, but it must be different from the ones already provided.

Problem: {question}

Existing solutions:
{solutions}

Use exactly this format (no extra text):
<New Solution> [Your reasoning should go here]. The answer is [answer]. </New Solution>
\end{Verbatim}
\vspace{-3mm}
\end{framed}

\section{Averaged Distinct N-gram Diversity}\label{app:diversity_formula}
Given a set of responses $S = \{y^{(i)}\}_{i=1}^{n}$, for $N \in \{1,\ldots,5\}$, we calculate the averaged distinct N-gram diversity for each set as: 
\[\text{avg. dist. N-gram} \, (S) = \sum_{N=1}^{5} \frac{ |\text{set}(\text{N-gram}(R_C))|}{|\text{N-gram}(R_C)|}.\]
The diversity metric is calculated as the mean avg. distinct N-gram diversity over the sets of responses.

\section{Prompt for Computation Flow Parsing}\label{app:comp_flow_prompt}
\begin{framed}
\vspace{-3mm}
\begin{Verbatim}[breaklines=true, 
                 breakanywhere=true,
                 breaksymbolright={},
                 breaksymbolsepleft=0pt,
                 breaksymbolindent=0pt,
                 breaksymbolleft={},
                 fontsize=\small]
You will receive a math question and a free-form solution. Extract the sequence of arithmetic steps from the solution and output them one by one.

Rules:
- Output ONLY lines made of digits 0-9, parentheses (), the operators + - * / ^, and optionally "=" to show each step's result.
- No words, units, currency symbols, or extra text.
- One step per line, in the order implied by the solution.
- Convert verbal quantities to numbers. Replace references like "the remainder" with the actual numeric value.
- Keep only the steps that lead to the final answer.
- If no computable arithmetic appears, output an empty line.

Example:

Question: Janet lays 16 eggs a day. She eats 3, uses 4 for baking, and sells the rest for $2 each. How much money does she make?
Solution: Janet lays 16 eggs per day. She eats 3 and uses 4 for baking, so 16 - 7 = 9 eggs left. She sells them at $2 each → 9 * 2 = $18.
Output:
3 + 4 = 7
16 - 7 = 9
9 * 2 = 18

Now, extract the arithmetic steps from the following:

Question: {question}
Solution: {solution}
Output:
\end{Verbatim}
\vspace{-3mm}
\end{framed}

\end{document}